\documentclass[conference]{IEEEtran}
\IEEEoverridecommandlockouts
\usepackage{cite}
\usepackage{multirow}
\usepackage{amsmath,amssymb,amsfonts}
\usepackage{algorithm, algpseudocode}
\usepackage{graphicx}
\usepackage{textcomp}
\usepackage{xcolor}
\usepackage{algpseudocode}
\usepackage{url}
\def\BibTeX{{\rm B\kern-.05em{\sc i\kern-.025em b}\kern-.08em
    T\kern-.1667em\lower.7ex\hbox{E}\kern-.125emX}}
\begin{document}

\title{A Genetic Algorithm with Tree-structured Mutation for Hyperparameter Optimisation of Graph Neural Networks
}

\author{\IEEEauthorblockN{Yingfang Yuan}
\IEEEauthorblockA{\textit{School of Mathematical} \\ \textit{and Computer Sciences} \\
\textit{Heriot-Watt University}\\
Edinburgh, United Kingdom \\
yyy2@hw.ac.uk}
\and
\IEEEauthorblockN{Wenjun Wang}
\IEEEauthorblockA{\textit{School of Mathematical} \\ \textit{and Computer Sciences} \\
\textit{Heriot-Watt University}\\
Edinburgh, United Kingdom \\
wenjun.wang@hw.ac.uk}
\and
\IEEEauthorblockN{Wei Pang{\footnotesize \textsuperscript{*}} \thanks{* Corresponding author}}
\IEEEauthorblockA{\textit{School of Mathematical} \\ \textit{and Computer Sciences} \\
\textit{Heriot-Watt University}\\
Edinburgh, United Kingdom \\
w.pang@hw.ac.uk}

}

\IEEEoverridecommandlockouts
\IEEEpubid{\makebox[\columnwidth]{978-1-7281-8393-0/21/\$31.00~\copyright2021 IEEE \hfill} 
\hspace{\columnsep}\makebox[\columnwidth]{ }}

\maketitle
\IEEEpubidadjcol
\begin{abstract}
In recent years, graph neural networks (GNNs) have gained increasing attention, as they possess the excellent capability of processing graph-related problems. In practice, hyperparameter optimisation (HPO) is critical for GNNs to achieve satisfactory results, but this process is costly because the evaluations of different hyperparameter settings require excessively training many GNNs. Many approaches have been proposed for HPO, which aims to identify promising hyperparameters efficiently. In particular, the genetic algorithm (GA) for HPO has been explored, which treats GNNs as a black-box model, of which only the outputs can be observed given a set of hyperparameters. However, because GNN models are sophisticated and the evaluations of hyperparameters on GNNs are expensive, GA requires advanced techniques to balance the exploration and exploitation of the search and make the optimisation more effective given limited computational resources. Therefore, we proposed a tree-structured mutation strategy for GA to alleviate this issue. Meanwhile, we reviewed the recent HPO works, which gives room for the idea of tree-structure to develop, and we hope our approach can further improve these HPO methods in the future.

\end{abstract}


\section{Introduction}
Graph is a data type used to describe the structured objects consisting of nodes and edges, such as citation networks, molecules, and road networks. In graph-related learning and prediction problems, traditional machine learning methods require using feature engineering to process graph data before training or prediction~\cite{zhou2018graph}. In contrast, graph neural networks (GNNs) are proposed to directly operate on graphs to solve graph-related problems in an end-to-end manner \cite{zhou2018graph}. GNNs exploit neural networks to model graphs and perform representation learning, which requires that GNNs are set with appropriate hyperparameters (e.g., learning rate, the number of neural network layers) to control the learning process. From our perspective, more sophisticated neural architectures are required to process graph data, which may result in challenging HPO tasks for GNNs. Falkner et al. \cite{falkner2018bohb} pointed out that deep learning algorithms are very sensitive to many hyperparameters. The work presented in \cite{yuan2021novel} demonstrated that HPO for GNNs is vital for achieving satisfactory results in practice. Therefore, research on effective HPO approaches for GNNs is critical and of great value for GNN applied to various real-world problems.

In brief, HPO is a trial-and-error process to obtain optimal solutions via iteratively generating and evaluating hyperparameter settings until the preset stopping condition(s) are satisfied. Commonly, the objective function of GNNs is selected as the fitness function of HPO to evaluate hyperparameters. However, the evaluation is often very expensive because training GNN models would take a lot of computational resources. On the other hand, with the development of deep learning, neural networks have an increasing number of layers or various other optional hyperparameters (e.g., activation functions, optimisation methods), which results in a large search space and increasing the difficulty of HPO. So, existing works of HPO have been conducted based on the issues of how to find the best solutions with fewer trials, reducing evaluation cost, and improving the capability of exploration. In this paper, a trial denotes a complete process of evaluating a hyperparameter setting (solution) on the fitness function (objective function).

There are many state-of-the-art HPO approaches with different search strategies. For example, TPE~\cite{bergstra2011algorithms} and BOHB~\cite{falkner2018bohb} are proposed based on Bayesian optimisation (BO) which are suitable for the functions that are expensive to evaluate. However, the former uses density functions to model two sets (good and bad) of hyperparameters, instead of using Gaussian process as in \cite{snoek2012practical, martinez2014bayesopt} as a standard surrogate model to approximate the distribution over the objective function. Furthermore, BOHB combines Bayesian optimisation and bandit-based methods to ensure good performance at anytime and speed up the process towards optimal solutions. In general, BO-based methods balance the exploration and exploitation by acquisition function. On the contrary, CMA-ES \cite{hansen2016cma} and HESGA \cite{yuan2021novel} are evolutionary approaches that exploit high-quality individuals to guide the generation of promising offspring, while they rely on evolutionary operators to explore search space. CMA-ES employs an adaptive multivariate distribution to sample individuals, and HESGA uses a genetic algorithm that relies on crossover and mutation operators to generate better individuals. To deepen the work of HESGA \cite{yuan2021novel}, our research focuses on improving the exploration capability of the mutation operator.

Genetic algorithm with hierarchical evaluation strategy (HESGA) \cite{yuan2021novel} utilises the elite archive to preserve excellent individuals, from which a parent can be selected to generate offspring. Meanwhile, the fast evaluation strategy has been introduced in HESGA to speed up optimisation and save computational resources. Furthermore, HSEGA uses the single point mutation as in \cite{deb1995simulated, lim2017crossover, whitley1994genetic} which is one factor for preventing GA from getting stuck in local optima and maintains the diversity of population \cite{mirjalili2019genetic}. However, we believe that the mutation operation in HSEGA can be further explored as only classical mutation operation was used in HSEGA. Therefore, in this research, we propose a tree-structured mutation (TSM) which employs the tree structure to store hierarchical historical information to guide mutation. In this way, GA can explore the whole search space more adaptively and effectively. Meanwhile, the idea of TSM is also compatible and scalable to other HPO algorithms that hierarchise historical information for future discovery.

Our contributions are summarised as follows:
\begin{itemize}
    \item We have developed a tree-structured mutation operation for GA to adaptively maintain the balance between exploration and exploitation for searching the hyperparameter space of GNNs while retaining the randomness of mutation operator.
    \item We conducted a review of advanced HPO methods, and we found that the tree-structured approach has the potential to be integrated into other approaches.
    \item Our research contributes to the development of molecular machine learning \cite{wu2018moleculenet} as well as HPO for GNNs in general.
    
    
\end{itemize}
The rest of this paper is organised as follows. Section II
introduces relevant work on HPO. In Section III, the details of our tree-structured mutation are presented. The experiments are reported and the results are analysed in Section IV. Finally, Section V concludes the paper and explores some directions in future work.

\section{Related Work}
In this section, we will first review some popular HPO methods. The genetic algorithm with hierarchical evaluation strategy (HESGA) \cite{yuan2021novel} will then be introduced because our tree-structured mutation is proposed upon it. Finally, some mutation strategies will be reviewed.

\subsection{Hyperparameter Optimisation Methods}
There have been a lot of state-of-the-art HPO methods, and these methods can be classified into \textit{black-box optimisation} and \textit{multi-fidelity optimisation} approaches \cite{elshawi2019automated}. The former focus on the effectiveness of the search algorithms and the latter focuses on efficiently using computational resources in search algorithms \cite{wu2020practical, yang2019conditional, hu2019multi}.


\textit{Black-box optimisation} \cite{golovin2017google, kandasamy2020tuning} implies that the details of the objective function and its gradient information is unknown during searching optimal solutions. In HPO, hyperparameters can be evaluated on the objective function, but we do not have access to any information about the model. Random search \cite{bergstra2012random} and grid search are two simple and commonly used methods for HPO. Bergstra et al. \cite{bergstra2011algorithms} hold the point that random search is more efficient than grid search given a fixed limited computation budget for HPO. However, facing the problems with expensive evaluations, it is very challenging for random search and grid search to achieve satisfactory performance. 

The use of \textit{Bayesian optimisation} (BO) can alleviate the issue of high computational cost~\cite{wu2020practical, kandasamy2020tuning}. The framework of BO is concluded in the sequential model-based algorithm \cite{hutter2011sequential} that iteratively exploits historical data to fit surrogate models or other transformations, while the most promising candidate is drawn according to a predefined criterion (i.e. acquisition function). The work presented in \cite{bergstra2011algorithms, martinez2014bayesopt, snoek2012practical, wu2019hyperparameter} makes use of Gaussian processes (GPs) to approximate the distribution over the objective function, and the candidate is selected by the acquisition function: probability of improvement (PI) or expected improvement (EI). In contrast, TPE \cite{bergstra2011algorithms} employs Parzen estimators as a surrogate to directly model promising and unpromising hyperparameters, and the candidate solutions are drawn according to EI that is proportional to the ratio of two density functions. During searching optimal solutions, the acquisition function plays a significant role in balancing exploration and exploitation.

Furthermore, \textit{evolutionary computation} has demonstrated the capability of solving HPO problems \cite{loshchilov2016cma, lorenzo2017hyper, hansen2016cma, yuan2021novel}. CMA-ES \cite{hansen2016cma} imitates the biological evolution, assuming that no matter what kind of gene changes, the results (traits) always follow a Gaussian distribution of a variance and zero-mean. Meanwhile, the generated population is evaluated on the objective function, and a portion of the individuals with excellent performance is selected to suggest evolution, moving to the area where better individuals would be drawn with higher probability. 

Successive halving \cite{jamieson2016non} is a bandit-based \textit{multi-fidelity} method for efficiently allocating computational resource that gives the most budget to the most promising individuals. Meanwhile, successive halving is an iterative process in which the best half of individuals are retained, and half of the individuals with lower performance are discarded when a portion of the total computational resource runs out. Hyperband \cite{li2017hyperband} and BOHB \cite{falkner2018bohb} are proposed based on the successive halving, but the former proposed a modification that frequently performs the successive halving method with different budget allocations to find the best solutions. Li et al. \cite{li2017hyperband} pointed out that because the terminal losses are unknown, the trials of different allocations are helpful to recognise high-quality solutions. For example, two individuals might not have a significant difference on fitness in the $30^{th}$ epoch, but a greater difference may be observed in the $50^{th}$ epoch if more computational resources (i.e., epochs) are allocated. Furthermore, BOHB combines Hyperband and BO to chase an ideal state in which performance and computational cost are both covered.

\subsection{Genetic Algorithm}
\textit{Genetic algorithm} (GA) is a class of evolutionary computation has been applied to HPO \cite{di2018genetic, han_hyperparameter_2020, yuan2021novel}. GA evaluates the qualities of individuals in the population by using a fitness function. To search for high-quality individuals, GA relies on crossover, mutation, and selection operators. The selection is inspired by natural selection, which means GA selects individuals as parents according to the probabilities (e.g., roulette wheel) to generate next-generation by crossover and mutation, and the probabilities are proportional to their quality (fitness). There are many crossover techniques such as uniform crossover~\cite{semenkin2012self}, single-point~\cite{srinivas1994genetic}, and heuristic crossover~\cite{grefenstette1985genetic}; they aim to recombine genes from the parents to generate offspring. In the evolutionary process, mutation operators as the last step are generally assigned with a low probability to be invoked. However, it plays a significant role for preventing GA from getting stuck in local optima. Similar to the probabilistic sampling of solutions in BO-based optimisation, the mutation is employed to endue GA with the exploration capability of solutions.

Variable length genetic algorithm \cite{xiao2020efficient} has been proposed to solve neural network HPO problem by increasing the length of the chromosome (i.e., enlarge search space) if the fitness does not satisfy preset conditions. Moreover, HESGA \cite{yuan2021novel} introduced the elite archive mechanism that stores some current optimal individuals. According to the roulette wheel, parents are selected from the elite archive and the population (Fig. \ref{fig:hesga}). In this way, new offspring always own at least a portion of the strong genes from the parent in the elite archive if crossover occurs. Meanwhile, HESGA provides the strategy of using fast and full evaluation methods, to alleviate the expensive evaluations. Fast evaluation is applied to the whole population. The individuals with the most difference of fitness in the early training stage (10\% $\sim$ 20\% of the total epochs) are considered promising, and it assumes that individuals with a steeper learning curve have more significant potential to achieve satisfactory results. Meanwhile, the full evaluation is employed to measure a small portion of individuals selected by fast evaluation. The fast and full evaluation can be considered \textit{multi-fidelity optimisation}, which focuses on decreasing the evaluation cost by combining many low cost evaluations with low-fidelity and a small number of costly evaluation with high-fidelity. 


\subsection{Mutation} \label{sec:mutation}
Many mutation strategies have been proposed for GA, and we only present some of them which are related to our research. Random (uniform) mutation and non-uniform mutation have been proposed in \cite{michalewicz2013genetic}. In random mutation, where a gene is replaced with a random value drawn from a defined range. On the other hand, in non-uniform mutation, the strength of mutation decreases along with the increasing number of generations. In adaptive mutation \cite{srinivas1994adaptive, whitley1990genitor}, each solution is assigned with a mutation rate to maintain the diversity of the population without affecting algorithm performance. Based on Gaussian mutation, self-adaptive mutation \cite{489178} allows GA to vary the mutation strength during the evolution, while self-adaptive Gaussian mutation based adaptation of population size is proposed in \cite{10.1007/3-540-61723-X_1006}. In Pointed directed (PoD) mutation \cite{berry2004pod}, each gene is tightly associated with a single bit that guides the direction of mutation which the gene may follow. Compared with Gaussian mutation, Deep et al. \cite{deep2007new} reported that PoD mutation is more advantageous in unconstrained benchmark problems. The above mentioned mutation operators exploit the historical information which is generated during evolution to adjust the mutation strategy dynamically, which inspired our TSM to improve the naive single point mutation used in \cite{yuan2021novel}.

\section{Method}
In this section, the design of tree-structured mutation (TSM) will be described. Meanwhile, we will introduce the use of TSM in HESGA.

\subsection{HESGA}\label{sec:mutatedhesga}

\begin{figure}
    \centering
    \includegraphics[scale=0.5]{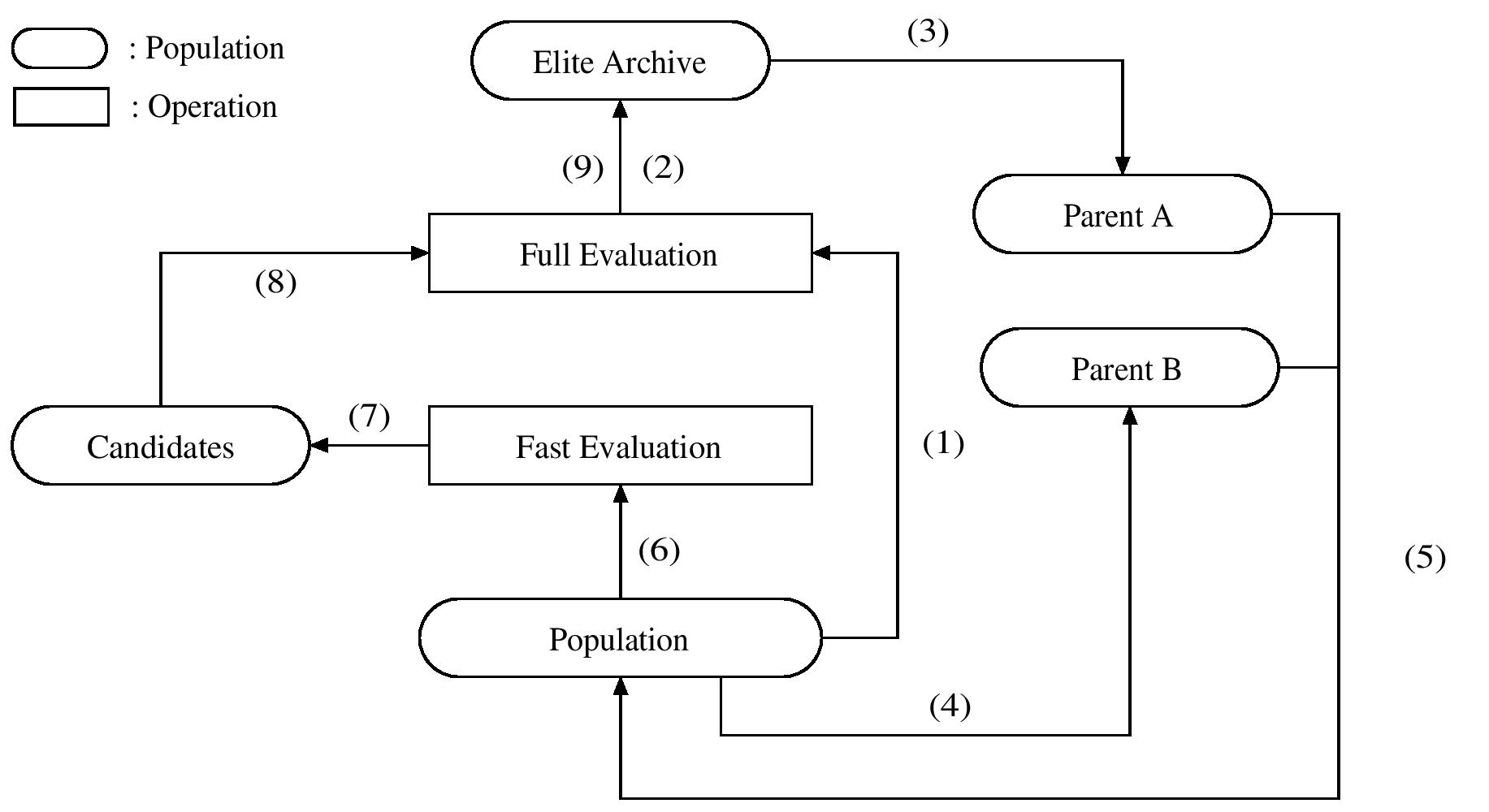}
    \caption{Genetic Algorithm with Hierarchical Evaluation Strategy \cite{yuan2021novel}}
    \label{fig:hesga}
\end{figure}

\begin{figure}
    \centering
    \includegraphics[scale=0.5]{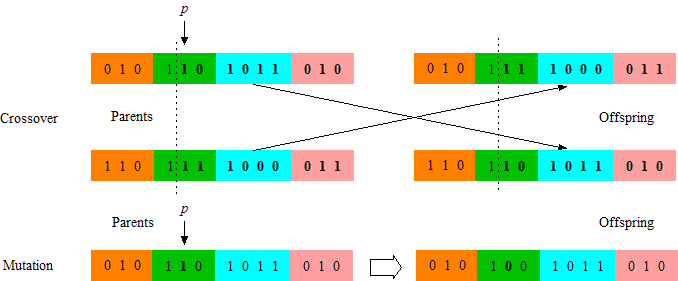}
    \caption{Binary Crossover and Mutation \cite{yuan2021novel}}
    \label{fig:crossoverandmutation}
\end{figure}

The framework of HESGA \cite{yuan2021novel} is shown in Fig. \ref{fig:hesga}. The TSM is proposed as a further improvement of HESGA. The original HESGA uses the binary encoding scheme for hyperparameters on HPO: each hyperparameter is assigned with a fixed-length binary code with a step size (i.e., resolution). For example, if the batch size ranges from 8 to 512 with a step size of 8, and then a batch size of 16 is represented by $[0,0,0,0,0,0,1]$. Single point crossover and mutation are employed in HESGA. Fig. \ref{fig:crossoverandmutation} shows that both the mutation and crossover are implemented by a uniform distribution to randomly generate a position $p$ at first. The bits on the right of the point $p$ are swapped between the two parent chromosomes for the crossover, and the bit on the position $p$ changes the value from 0 to 1, or vice versa for mutation. In initialisation, all randomly generated individuals are assessed by full evaluation, and some individuals with higher fitness values are selected to be included in the elite archive by Step (1) and (2) as shown in Fig. \ref{fig:hesga}. To update the elite archive, the offspring is generated by operating crossover and mutation on the two parents that are respectively selected by roulette wheel from the elite archive and population in Steps (3), (4) and (5). Then, in Step (6), fast evaluation is operated to find a small number of individuals (candidates) which have better fitness values, and these individuals are further accessed by full evaluation in Step (8). The elite archive is updated if candidates are better than some of the individuals in the elite archive in Step (9). The work presented in \cite{yuan2021novel} demonstrated that HESGA had achieved satisfactory performance on HPO for GNNs.

\subsection{Tree-structured Mutation}\label{sec:tsm}
\begin{figure}
    \centering
    \includegraphics[scale=0.5]{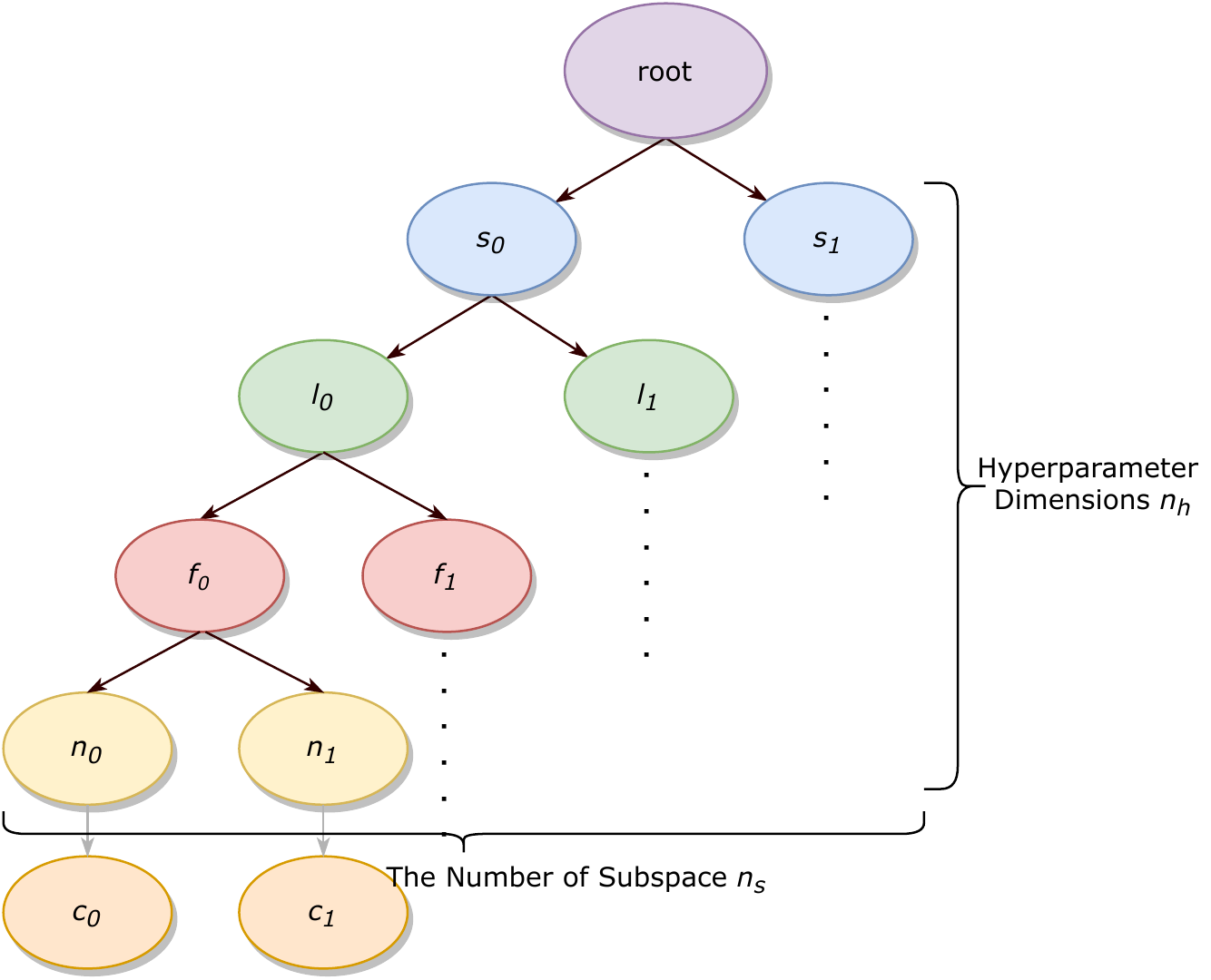}
    \caption{Tree-structured hyperparameter space, the internal nodes of $s, l, f, n$ represent the different hyperparameters and used to define their hyperparameter subspace with different subscripts, such as $s_{0}$ and $s_{1}$ respectively correspond to lower and upper portion of the hyperparameter $s$. $n_s$ denotes the maximum width of the tree and $n_h$ is the dimension of hyperparameter settings}
    \label{fig:treestrcuturedseachspace}
\end{figure}
In HPO using GA, the mutation operator is a way for GA to explore the hyperparameter search space. All mutation operators discussed in Section \ref{sec:mutation} are proposed based on the core principle of effective mutation towards optimal solutions. Most of the black-box optimisation methods utilise historical information to carry out effective exploration. Here, back to GA, we expect that the historical information can also be exploited with the tree to suggest mutations.

Considering that TPE \cite{bergstra2011algorithms} used tree structure to model the sampling of hyperparameter settings, we propose to employ a binary tree to build the hierarchical search space, as shown in Fig. \ref{fig:treestrcuturedseachspace}. Each pathway from the root node to a leaf node represents a subspace of the whole hyperparameter space. For example, from root to node $c_{0}$, the child nodes ($s_{0}, l_{0}, f_{0}, n_{0}$) define the corresponding ranges of hyperparameters \textit{(batch size, learning rate, the size of graph convolution layer, and the size of fully-connected layer)}, which forms a subspace of the whole search space and can be represented by a binary string $0000$. Meanwhile, the left and right child nodes define the lower and higher ranges of hyperparameters by thresholds. For example, if $s$ ranges from $8$ to $512$, the median value $256$ is chosen as the threshold to give two subspace with the same size. $s_{0}$ indicate a range from $8$ to $255$ and  $s_{1}$ ranges from $256$ to $512$. In this way, the whole search space is divided into many subspaces, and the number of subspaces equals the total number of leaf nodes $c$ (i.e., the maximum width ${n_{s}}$), or the square of the dimension of hyperparameter settings ${n_{h}}^{2}$, where $n_{h}$ is also the maximum depth of the tree without leaf nodes.

To use the tree to suggest mutations, the internal and leaf nodes take the responsibility to record historical information. The internal nodes are used to store the times of mutation that occurred on different dimensions in a subspace. Concurrently, a leaf node $c$ is used to record the times of a subspace that has been visited. For example, when a hyperparameter setting $\lambda$ sampled from the subspace ($s_{0}, l_{0}, f_{0}, n_{0}$) has been evaluated on an objective function, the value of node $c_{0}$ will increase by 1. If $\lambda$ underwent a mutation on \textit{batch size}, then the counter in the $s_{0}$ node will increase by 1. In this way, the whole search space is hierarchised which increase the ability of exploration for mutation. 

\begin{figure}
    \centering
    \includegraphics[scale=0.5]{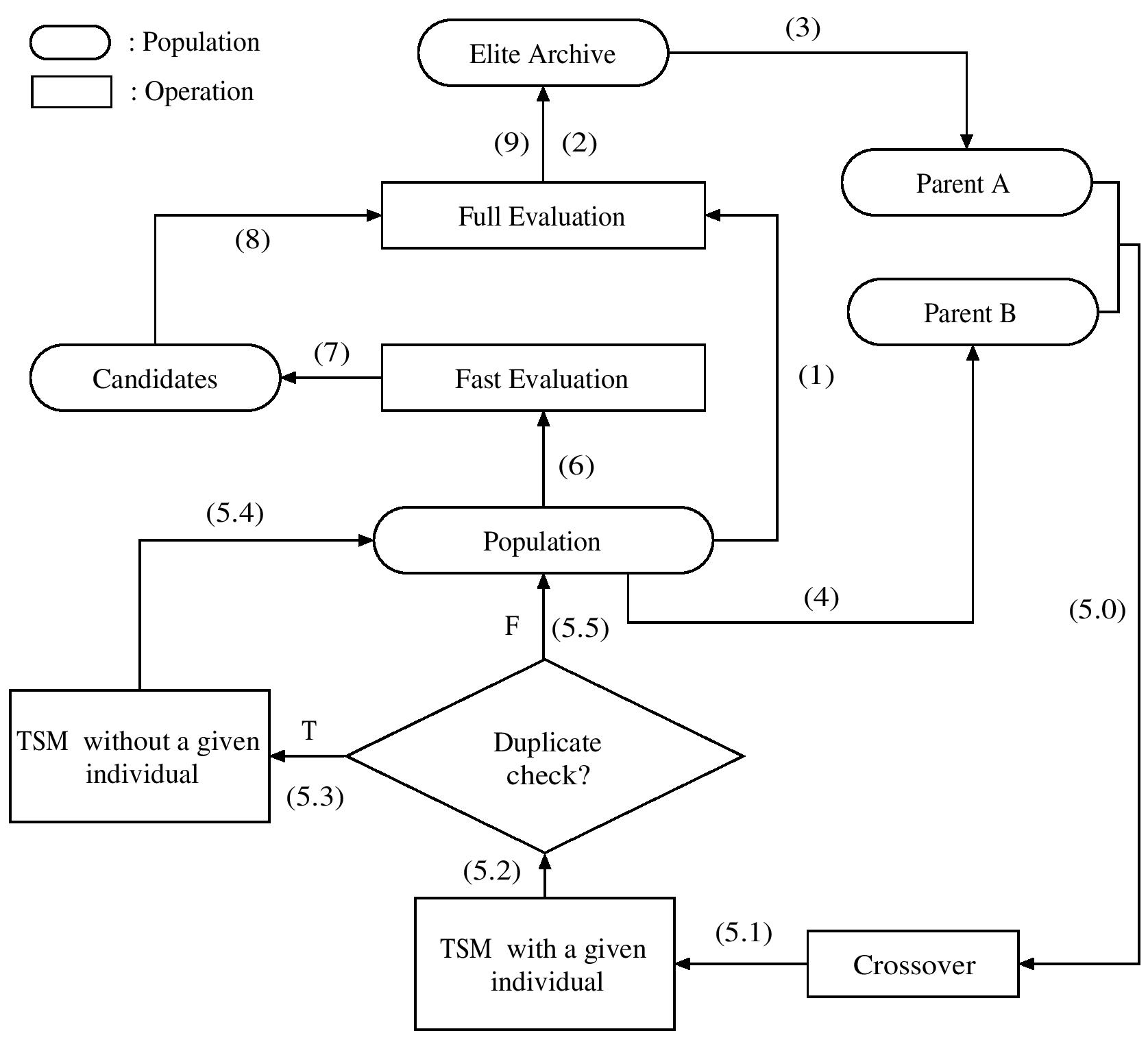}
    \caption{Genetic Algorithm with Hierarchical Evaluation Strategy with TSM mutation}
    \label{fig:tsmhesga}
\end{figure}

In this research, based on the framework of HESGA, we use TSM to replace the single point mutation, considering that a more sophisticated mutation may lead to better performance. We observed that there might exist the same individuals in elite archive and populations that waste computational resources for evaluating them during the experiments. Many reasons may lead to this issue. For example, an extremely excellent individual in the population will be selected to be included in the elite archive while outperforms other individuals in the elite archive; it means this individual has a higher probability of being selected as both Parent A and B in the next generation. Two identical parents will invalid crossover by which the same individuals will be generated. To solve this issue, we set the condition in Step (5.2) (Fig. \ref{fig:tsmhesga}) to prohibit duplicates. If the duplication is detected, the mutation is forcibly invoked. Based on TSM, we use two strategies: TSM with a given individual and TSM without a given individual respectively for the situation: mutation is invoked with preset probability in Steps (5.1 and 5.2) and force mutation when duplicates are detected in Steps (5.3 and 5.4), as shown in Fig. \ref{fig:tsmhesga}.

\subsubsection{TSM with A Given Individual} \label{tsmwithan}
HESGA operates the single point mutation with the probability of 0.2, as indicated by  Step (5) in Fig. \ref{fig:hesga}. TSM is also invoked with the same probability in Step (5.1) in Fig. \ref{fig:tsmhesga}. TSM uses historical information recorded in the tree to generate a probability distribution to guide the mutation compared with the change of a bit of binary encoding under a uniform distribution. Specifically, when mutation is activated, for the individual concerned, its corresponding subspace (i.e., pathway) along the tree will be identified. In that pathway, the times $t$ of previously occurred mutation respectively according to the mutated position stored in the internal nodes will be given to Equation (\ref{eq:rw}) to compute probabilities. In Equation (\ref{eq:rw}), $i$ is the depth of an internal node given pathway $p$, and $t_{i}$ represents the times recorded in that node. $f$ is a reciprocal function, $j$ are the depth of all internal nodes (e.g., $n_{1}$ with depth 4, $f_{0}$ with depth 3, $l_{0}$ with depth 2, and $s_{0}$ with depth 1 ) in pathway $p$, and $t_{j}$ is the times stored in those nodes respectively. The probabilities will determine where the mutation happens following the roulette wheel method, and the mutated individual will be used to update the tree to determine future mutation. In this way, in a subspace, the dimension that has been previously explored frequently will have less probability to be explored in the future. When the dimension is sampled, the new value is drawn from the search space defined in that node by the uniform distribution. For example, when $s_{0}$ is selected, a new value from 8 to 255 with the resolution of 8 will be drawn and converted into binary coding to replace the original fragment. The TSM with a given individual is summarised in Algorithm 1.

\begin{equation} \label{eq:rw}
    \mathcal{P}\left(t_{i}\right)=\frac{f\left(t_{i}\right)}{\sum_{j=1}^{n_{h}} f\left(t_{j}\right)} \ i,j \in  \{1,2,...,n_{h}\}, \text{given~} p.
\end{equation}

\begin{algorithm}[!]
\caption{TSM with A Given Individual}
\label{alg1}
\begin{algorithmic}[1]
    \State \textbf{Input} an individual $s$, $tree$, $t = [~]$
    \State $p$ = {$tree(s)$} \Comment{pathway $p$ identification in tree} 
    \For{$j = 1$ $\rightarrow$	~$n_{h}$ }  \Comment{in pathway $p$, the maximum depth of tree without leaf nodes $n_{h}$}
         \State  $t_{total} += f\left(t_{j}\right)$ \Comment{$t_{j}$ denotes the times recorded in node $j$}, $f$ is reciprocal function 
         \State $t$.append($f(t_{j})$)
    \EndFor
    \State $t^{\prime}$ $\leftarrow$ each element in $t$ is divided by $t_{total}$
    \State $i$ = $rws$($n_{h}$, $t^{\prime}$) \Comment{node $i$, $rws$ roulette wheel selection}
    \State $r \leftarrow$ the defined value range in node $i$
    \State $v = uniform(r)$ \Comment{mutated value $v$}
    \State $s^\prime$ $\leftarrow$ $s_{i}$ is replaced with $binary(v)$ \Comment{ for $s$, $s_{i}$ the fragment of binary coding for mutation}
\State \textbf{Output} individual $s^{\prime}$
\end{algorithmic}
\end{algorithm}

\subsubsection{TSM without A Given Individual}
To avoid duplicated individuals existing in the elite archive, in Fig. \ref{fig:hesga}, Step (9) is added with a duplication detection mechanism. The duplicates will be deleted except the one with the best fitness value (in our experiments, there exist fluctuations when evaluating the same hyperparameter settings due to the randomness in the GNN training and evaluation process). Meanwhile, if the duplicates are found in the population, we will take TSM without a given individual (shown in Fig. \ref{fig:tsmhesga}, Steps (5.3 and 5.4)) to generate a new individual to replace them. TSM with a given individual is also feasible to deal with the duplication issue, but here we proposed the TSM without a given individual to increase the population diversity.

Similar to Algorithm 1, TSM without a given individual (Algorithm 2) also exploits the information stored in the tree and follows the roulette wheel to select mutation points. However, the latter selects a subspace to sample new individuals (mutated ones) rather than given a subspace to select a dimension to mutate. Practically, the mutation starts by computing the probabilities of each subspace by Equation \ref{eq:rwfull}, where $c$ are the indexes of leaf nodes, $n_{s}$ is the total number of leaf nodes (i.e., maximum width), $f$ is the reciprocal function. Each leaf node corresponds to a subspace, and the times of a subspace being explored is recorded on it. Eq. (\ref{eq:rwfull}) indicates that the subspace with a large number of $t$ will have less probability to be explored by mutation. In this way, exploitation can be achieved by crossover and elite archive, and exploration is facilitated by mutations. The pseudo-code is shown in Algorithm 2.

\begin{equation} \label{eq:rwfull}
    \mathcal{P}\left(t_{i}\right)=\frac{f\left(t_{i}\right)}{\sum_{c=1}^{n_{s}} f\left(t_{c}\right)} \  i,c \in \{1,2,3...n_{s}\}, \text{given leaf nodes}. 
\end{equation}

\begin{algorithm}[!]
\caption{TSM without A Given Individual}
\label{alg2}
\begin{algorithmic}[1]
    \State \textbf{Input} $tree$, $t=[~]$
    \For{$c = 1$ $\rightarrow$	~$n_{s}$ } 
         \State  $t_{total} += f\left(t_{c}\right)$ \Comment{$t_{c}$ denotes the times recorded in node $c$, $f$ is reciprocal function} 
         \State $t$.append($f(t_{c})$)
    \EndFor
    \State $t^{\prime}$ $\leftarrow$ each element in $t$ is divided by $t_{total}$
    \State $i$ = roulette wheel($n_{s}$, $t^{\prime}$) \Comment{subspace $i$}
    \State $v = [~]$
    \For{$j = 1$ $\rightarrow$	~$n_{h}$ } \Comment{in subspace $i$}
        \State $r_{j} \leftarrow$ the defined value range in node with depth $j$
        \State $v_{j} = uniform(r_j)$ \Comment{mutated value $v_{j}$ by sampling from a uniform distribution given range $r_{j}$}
        \State $v$.append$(binary(v_{j}))$
    \EndFor
    \State $s$ = $v$
\State \textbf{Output} individual $s$
\end{algorithmic}
\end{algorithm}

\section{Experiments}
To assess the effectiveness of TSM, we conducted the experiments to compare HESGA with single-point mutation \cite{yuan2021novel} and HESGA equipped with TSM. The HESGA in our experiments is modified to reject duplication, and the duplicates found in the population will be forcibly re-sampled by TSM without a given individual (as mentioned in Section \ref{sec:tsm}) or single-point mutation. The settings of HESGA are as follows: the max generation is set to 10, the probabilities of crossover and mutation are 0.8 and 0.2, which are the same as the original experimental settings \cite{yuan2021novel}. Meanwhile, the proportion of the size of elite archive to the population size is 0.5, as considering that 0.1 may lead to the selection pressure. 

Our experiments have been conducted on three benchmark datasets: ESOL, FreeSolv, and Lipophilicity from \cite{wu2018moleculenet}. They are three molecular property prediction datasets that are used to assess the performance of GNNs. Meanwhile, there are many types of GNN algorithms, but the neural fingerprint graph model (GC) \cite{duvenaud2015convolutional} has been selected because it was proposed to be used for molecular machine learning. GC exploits the neural architectures to model circular fingerprint \cite{glen2006circular} for learning the molecular representations for prediction tasks. There are four hyperparameters selected for HPO, including $s_b$ (batch size), $l_r$ (learning rate), $s_g$ (the size of graph convolution layer), and $s_f$ (the size of fully-connect layer); the search space is constructed as follows: the range of $s_b$ is $8\sim512$ with step size 8, the range of $l_r$ is $0.0001\sim0.0032$ with step size 0.0001, the range of $s_g$ is 8 $\sim$ 512 with step size 8, and the range of $s_f$ is 32 $\sim$ 1024 with step size 32. As mentioned in Section \ref{sec:tsm}, TSM requires thresholds to divide the whole search space into many subspace, and we uses the median values of $s_b$, $l_r$, $s_g$, and $s_f$, which are 256, 0.0016,  256, and 512, respectively to build the binary tree in experiments.

As in \cite{yuan2021novel}, the root mean square error (RMSE) of GC is used as the fitness function for HPO, and it is also used to measure the HPO performance. The best hyperparameter setting found by an HPO method (HESGA and HESGA+TMS) will be given to GC to run 30 times, and the mean of the 30 RMSEs are used to measure the performance of HPO. Meanwhile, in order to make the comparisons more convincing, $t$-test with a significance level of two tails 10\% is employed to compare the performance of HESGA and HESGA + TSM and see if there exists a significant difference. In $t$-test, negative $t$ value and the hypothesis $h = 1$ indicates that HESGA outperforms HESGA+TSM; positive $t$ value and the hypothesis $h = 1$ indicates that HESGA+TSM significantly outperform HESGA; or $h = 0$ (the null hypothesis) means that there is no significant difference between HESGA and HESGA+TSM.

In the experiments on ESOL (Table \ref{tab:esol}), HESGA+TSM has a smaller mean and standard deviation of RMSEs than HESGA. However, there is no significant difference between HESGA and HESGA+TSM at the significance level $\alpha = 10\%$. We considered that the size of the ESOL dataset constrains the performance of HESGA+TSM, because a sufficient amount of data is a prerequisite for GC possessing better performance with a more complex architecture. In Table  \ref{tab:esol}, HESGA+TSM found larger $s_g$ and $s_f$, which means the GC with more parameters, and they require more training data to tune them for better performance. Furthermore, if we set the significance level $\alpha = 20\%$, HESGA+TSM outperforms HESGA in the validation data set. Meanwhile, it is obvious that two hyperparameter settings are completely different and come from different two subspaces, which indicates the complex and multi-modal nature of the underlying hyperparameter search space, and the discovery in different subspaces is meaningful. 

In the experiments on FreeSolv (Table \ref{tab:freesolv}), HESGA presents better performance than HESGA+TSM. FreeSolv is the smallest dataset used in our experiments. The small size of the dataset is intrinsically challenging for GC to learn and keep a stable performance, which may lead to the underperformance of HESGA+TSM since its strength is exploration, stable and accurate feedback (i.e., the results of evaluation on GC) is essential for conducting efficient exploration. 

Furthermore, when duplicates appear in the population, HESGA uses single-point mutation which retains most of the genes from parents, and relatively speaking the strength of the mutation is small, while HESGA+TMS is guided by the tree-based mutation, which explores the search space more broadly compared with HESGA. Therefore, we considered that when the size of a dataset is relatively small, global exploration may bring more fluctuation and unknown factors upon the instability of GC, which brings more difficulties to the problems. In contrast  ``local" exploration might be more effective since it reduces the underlying unknown risks from the global area.

In contrast, Lipophilicity is the largest dataset in our experiments, and as shown in Table \ref{tab:lipo}, HESGA+TSM outperforms HESGA in this dataset with a significant difference, and lower standard deviation means the result of HESGA+TSM is more stable. When the size of dataset increased, the full evaluation becomes more accurate, and it provides more helpful information for GA to generate strong offspring in terms of exploitation. Meanwhile, more effective exploration becomes necessary, and HESGA+TSM shows better performance because TSM uses historical information to search hyperparameter space rather than search each part of the hyperparameter space with equal probability.

\begin{table*}[!]
\centering
\caption{The Results on ESOL Dataset}
\label{tab:esol}
\begin{tabular}{|l|l|l|l|l|l|} 
\hline
ESOL                                                                    & Hyperparameters                                      & Train                   & Validation               & Test                    &                             \\ 
\hline
\multirow{4}{*}{HESGA}                                                  & $s_{b}$=64                                           & \multirow{2}{*}{0.2847} & \multirow{2}{*}{0.8843}  & \multirow{2}{*}{0.8862} & \multirow{2}{*}{Mean RMSE}  \\ 
\cline{2-2}
                                                                        & $l_{r}$=0.0028                                       &                         &                          &                         &                             \\ 
\cline{2-6}
                                                                        & $s_g$=304                                          & \multirow{2}{*}{0.0424} & \multirow{2}{*}{0.0509}  & \multirow{2}{*}{0.0506} & \multirow{2}{*}{Mean STD}   \\ 
\cline{2-2}
                                                                        & $s_f$=64                                           &                         &                          &                         &                             \\ 
\hline
\multirow{4}{*}{\begin{tabular}[c]{@{}l@{}}HESGA\\ + TSM \end{tabular}} & $s_{b}$=48                                           & \multirow{2}{*}{0.2734} & \multirow{2}{*}{\textbf{0.86585}} & \multirow{2}{*}{\textbf{0.8743}} & \multirow{2}{*}{Mean RMSE}  \\ 
\cline{2-2}
                                                                        & $l_{r}$=0.0012                                       &                         &                          &                         &                             \\ 
\cline{2-6}
                                                                        & $s_g$=320                                          & \multirow{2}{*}{0.0333} & \multirow{2}{*}{0.0357}  & \multirow{2}{*}{0.0463} & \multirow{2}{*}{Mean STD}   \\ 
\cline{2-2}
                                                                        & $s_f$=320                                          &                         &                          &                         &                             \\ 
\hline
\multicolumn{2}{|l|}{\begin{tabular}[c]{@{}l@{}}$t$-test on results\\ with significance level of $\alpha = 10\%$ \end{tabular}}   & $t=1.1249,~h=0~~$          & $t=1.5963,~h=0~~$           & $t=0.9311,~h=0~~$          &                             \\
\hline
\end{tabular}
\end{table*}

\begin{table*}[!]
\centering
\caption{The Results on FreeSolv Dataset}
\label{tab:freesolv}
\begin{tabular}{|l|l|l|l|l|l|} 
\hline
FreeSolv                                                                & Hyperparameters                                      & Train                   & Validation              & Test                    &                             \\ 
\hline
\multirow{4}{*}{HESGA}                                                  & $s_{b}$=24                                           & \multirow{2}{*}{0.5498} & \multirow{2}{*}{\textbf{1.0970}} & \multirow{2}{*}{\textbf{1.0370}} & \multirow{2}{*}{Mean RMSE}  \\ 
\cline{2-2}
                                                                        & $l_{r}$=0.0030                                       &                         &                         &                         &                             \\ 
\cline{2-6}
                                                                        & $s_g$=208                                          & \multirow{2}{*}{0.1251} & \multirow{2}{*}{0.1199} & \multirow{2}{*}{0.1246} & \multirow{2}{*}{Mean STD}   \\ 
\cline{2-2}
                                                                        & $s_f$=224                                          &                         &                         &                         &                             \\ 
\hline
\multirow{4}{*}{\begin{tabular}[c]{@{}l@{}}HESGA\\ + TSM \end{tabular}} & $s_{b}$=8                                            & \multirow{2}{*}{1.1181} & \multirow{2}{*}{1.4083} & \multirow{2}{*}{1.3508} & \multirow{2}{*}{Mean RMSE}  \\ 
\cline{2-2}
                                                                        & $l_{r}$=0.0010                                       &                         &                         &                         &                             \\ 
\cline{2-6}
                                                                        & $s_g$=480                                          & \multirow{2}{*}{0.2692} & \multirow{2}{*}{0.2408} & \multirow{2}{*}{0.2874} & \multirow{2}{*}{Mean STD}   \\ 
\cline{2-2}
                                                                        & $s_f$=576                                          &                         &                         &                         &                             \\ 
\hline
\multicolumn{2}{|l|}{\begin{tabular}[c]{@{}l@{}}$t$-test on results\\ with significance level of $\alpha = 10\%$ \end{tabular}}   & $t=-10.3063, h=1$           & $t=-6.2298, h=1$                 & $t=-5.3940, h=1$                 &                             \\
\hline
\end{tabular}
\end{table*}

\begin{table*}[!]
\centering
\caption{The Results on Lipophilicity Dataset}
\label{tab:lipo}
\begin{tabular}{|l|l|l|l|l|l|} 
\hline
Lipophilicity                                                                    & Hyperparameters                                      & Train                   & Validation              & Test                    &                             \\ 
\hline
\multirow{4}{*}{HESGA}                                                  & $s_{b}$=136                                          & \multirow{2}{*}{0.3656} & \multirow{2}{*}{0.7402} & \multirow{2}{*}{0.7293} & \multirow{2}{*}{Mean RMSE}  \\ 
\cline{2-2}
                                                                        & $l_{r}$=0.0031                                       &                         &                         &                         &                             \\ 
\cline{2-6}
                                                                        & $s_g$=232                                          & \multirow{2}{*}{0.0881} & \multirow{2}{*}{0.0476} & \multirow{2}{*}{0.0462} & \multirow{2}{*}{Mean STD}   \\ 
\cline{2-2}
                                                                        & $s_f$=1024                                         &                         &                         &                         &                             \\ 
\hline
\multirow{4}{*}{\begin{tabular}[c]{@{}l@{}}HESGA\\ + TSM \end{tabular}} & $s_{b}$=168                                          & \multirow{2}{*}{0.2390} & \multirow{2}{*}{\textbf{0.6962}} & \multirow{2}{*}{\textbf{0.7021}} & \multirow{2}{*}{Mean RMSE}  \\ 
\cline{2-2}
                                                                        & $l_{r}$=0.0011                                       &                         &                         &                         &                             \\ 
\cline{2-6}
                                                                        & $s_g$=304                                          & \multirow{2}{*}{0.0676} & \multirow{2}{*}{0.0284} & \multirow{2}{*}{0.0269} & \multirow{2}{*}{Mean STD}   \\ 
\cline{2-2}
                                                                        & $s_f$=864                                          &                         &                         &                         &                             \\ 
\hline
\multicolumn{2}{|l|}{\begin{tabular}[c]{@{}l@{}}$t$-test on results\\ with significance level of $\alpha = 10\%$ \end{tabular}}   & $t= 6.1358,~ h=1~~$        & $t=4.2642,~ h=1~~$            & $t=2.7381,~ h=1~~$            &                             \\
\hline
\end{tabular}
\end{table*}



\section{Conclusion and Future Work}
In this paper, we proposed TSM for genetic algorithm as applied to HPO of GNNs, which improved the performance of HESGA on the largest dataset in our experiments. Compared with operating HPO on a relatively small dataset, the achieved improvement of HPO performance on the larger dataset is more meaningful since the evaluations are more expensive. Meanwhile, the research of GA-based HPO for GNN is in an early stage; our work will help other researchers to better explore this area. On the other hand, our research contributes to the development of molecular machine learning, which may facilitate the research from the relevant domain such as materials science, biology and drug discovery. Furthermore, the idea of utilising tree structure to hierarchise search space and record the historical information can be applied to other HPO methods. Finally, in future, we believe more research can be done along further developing TSM for evolutionary HPO. We list some potential directions as follows.

\subsection{Reward and Rejection Mechanism}
Currently, the inner and leaf nodes are used to store the times of previous exploration. Inspired by reinforcement learning \cite{van2017hybrid}, the results of evaluations on fitness function can be recorded as rewards with the times together to affect future discovery by reconstructing Eqs. (1) and (2), and in this way, TSM is not only for increasing the exploration ability, but also for improving the exploitation in the later stage by introducing a reward mechanism. Meanwhile, the times and rewards can also work together in a surrogate model to reject those unpromising trials according to historical information. 

\subsection{Adaptive Threshold}
In TSM, the binary tree requires thresholds to be specified to generate child nodes. In our experiments, we choose median values as thresholds. However, it is easy to observe that $s_b$ in all experiments are from the lower range (i.e. the left child node). If the thresholds are adaptive and can be updated by the rewards referring to the percentile in TPE \cite{bergstra2011algorithms}, TSM may be further improved to better focus on adaptively adjusted subspace.

\section*{ACKNOWLEDGMENTS}
This research is supported by the Engineering and Physical Sciences Research Council (EPSRC) funded Project on New Industrial Systems: Manufacturing Immortality (EP/R020957/1). The authors are also grateful to the Manufacturing Immortality consortium.

\section*{Data Statement}
All data used in our experiments are from MoleculeNet \cite{wu2018moleculenet}, which are publicly available in \url{http://moleculenet.ai/datasets-1}.



\bibliographystyle{IEEEtran}
\bibliography{IEEEabrv,mybib}

\end{document}